\documentclass[conference]{IEEEtran}

\usepackage{times}
\usepackage{graphicx}
\usepackage{latexsym}
\usepackage{color}
\usepackage{amsmath}
\usepackage{setspace}
\usepackage{multirow}
\usepackage{url}

\usepackage{cite}
\usepackage{amsmath,amssymb,amsfonts}
\usepackage{algorithmic}
\usepackage{graphicx}
\usepackage{textcomp}
\usepackage{xcolor}

\begin{document}

\title{Simple and Lightweight Human Pose Estimation}

\author{
    \textbf{Zhe Zhang}, \ \textbf{Jie Tang} and \textbf{Gangshan Wu} \\
    \textit{State Key Laboratory for Novel Software Technology, Nanjing University, China} \\
    {\tt\small zhangzhe@smail.nju.edu.cn, \{tangjie,gswu\}@nju.edu.cn} 
}

\maketitle
\bibliographystyle{IEEEtran}

\begin{abstract}

Recent research on human pose estimation has achieved significant improvement. However, most existing methods tend to pursue higher scores using complex architecture or computationally expensive models on benchmark datasets, ignoring the deployment costs in practice. In this paper, we investigate the problem of simple and lightweight human pose estimation.

We first redesign a lightweight bottleneck block with two concepts: \textit{depthwise convolution} and \textit{attention mechanism}. And then, based on the lightweight block, we present a Lightweight Pose Network (LPN) following the architecture design principles of SimpleBaseline \cite{xiao2018simple}. The model size (\#Params) of our small network LPN-50 is only 9\% of SimpleBaseline(ResNet50), and the computational complexity (FLOPs) is only 11\%. We also propose an iterative training strategy and a model-agnostic post-processing function \textit{$\beta$-Soft-Argmax} to give full play to the potential of our LPN and get more accurate predicted results. We empirically demonstrate the effectiveness and efficiency of our methods on the benchmark dataset: the COCO keypoint detection dataset. Besides, we show the speed superiority of our lightweight network at inference time on a non-GPU platform. Specifically, our LPN-50 can achieve 68.7 in AP score on the COCO test-dev set, with only 2.7M parameters and 1.0 GFLOPs, while the inference speed is 17 FPS on an Intel i7-8700K CPU machine. The code is available in this URL\footnote{https://github.com/zhang943/lpn-pytorch}.

\end{abstract}

\section{Introduction}

Human pose estimation (HPE) is the problem of locating body keypoints (elbows, wrists, knees, etc.) from input images. It is a fundamental task in computer vision and has several practical applications such as action recognition \cite{wang2013approach,cheron2015p}, tracking \cite{cho2013adaptive}, and human-computer interaction \cite{shotton2011real}. Recently, significant improvements on this topic have been achieved by using deep convolutional neural networks (DCNNs) \cite{newell2016stacked,chen2017adversarial,chu2017multi,ke2018multi,xiao2018simple,li2019rethinking,sun2019deep}.

However, these state-of-the-art methods usually involve very wide and deep networks, with numerous parameters and a huge number of floating-point operations (FLOPs). Despite top-performing, one major drawback of such complex models is that they are very time-consuming at inference time because of the heavy computation. Moreover, having large amounts of parameters makes the models high memory demanding. For these reasons, it is not suitable to deploy the trained top-performing networks directly on resource-limited devices such as smartphones and robots. In the meantime, the demand for human pose estimation networks with small model size, light computation cost, and high accuracy is increasing. Bulat et al. \cite{bulat2017binarized} attempted to binarize the network architecture for model compression and execution speedup, which however suffered performance drop significantly. There is still insufficient research on lightweight human pose estimation.

\begin{figure}[t]
  \centering
  \includegraphics[width=\linewidth]{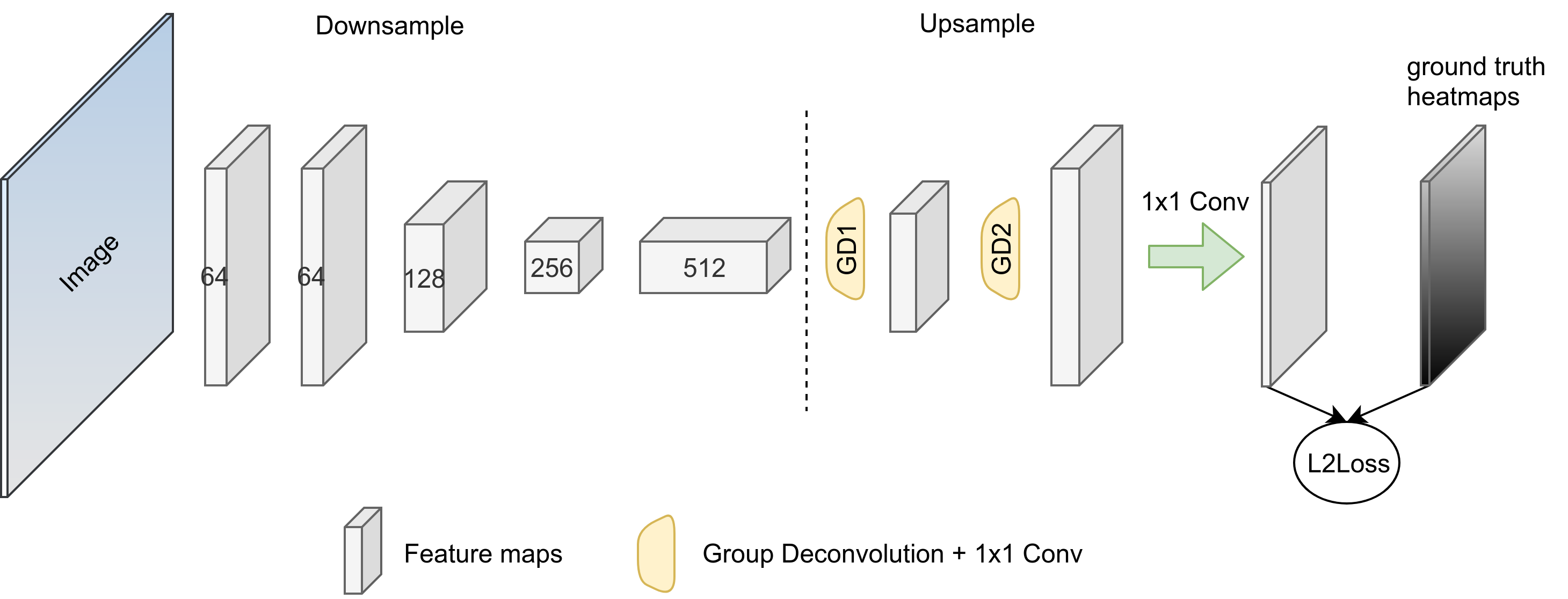}
  \caption{Illustrating the architecture of the presented LPN. Similar to SimpleBaseline \cite{xiao2018simple}, our LPN consists of a backbone network and several upsampling layers. However, differently, we use the redesigned lightweight bottleneck block as the basic component while downsampling, and we also choose a lightweight fashion for upsampling. See Section \ref{section:LPN} for details.}
  \label{fig:network}
\end{figure}

The success of SimpleBaseline \cite{xiao2018simple} has provided prior knowledge on how to design a simple network for human pose estimation and shown how good could a simple method be. Inspired by their graceful design, in this work, we focus on the problem of simple and lightweight human pose estimation. To achieve this purpose, we first analyze the parameter composition of the bottleneck block \cite{he2016deep}, which is the basic component of SimpleBaseline \cite{xiao2018simple}. And then, we redesign a lightweight bottleneck block that mainly exploits the best current design choice \textit{depthwise convolution} for network architecture with a low memory footprint. Besides, we also attempt to use a new \textit{attention mechanism} proposed by Cao et al. \cite{cao2019GCNet} to improve the capacity of the lightweight block. We show that the lightweight block is a simple drop-in replacement of the standard bottleneck block, but it can reduce the model size and computational complexity significantly without too much performance degradation. To further demonstrate the effectiveness and efficiency of the lightweight block, we present a Lightweight Pose Network (LPN) following the architecture design principles of SimpleBaseline \cite{xiao2018simple}. The architecture of the network is illustrated in Figure {\color{red} \ref{fig:network}}.

Human pose estimation networks rely heavily on pre-trained models. The experiments in \cite{xiao2018simple,sun2019deep} have shown that the performance of the network initialized with a pre-trained model is better than the network trained from scratch. However, the pre-trained model of SimpleBaseline \cite{xiao2018simple} can no longer be utilized since the network architecture has been changed after applying the lightweight bottleneck block. No doubt that training the network on ImageNet to get a pre-trained model will increase the time costs. To overcome this barrier, we propose an iterative training strategy. The training strategy changes the learning rate periodically so that it can give full play to the potential of our lightweight network. Our training strategy can achieve more significant improvement than using a pre-trained model.

Besides, we observe that almost all existing methods use function \textit{argmax} to get the maximum position in heatmaps and then calculate the final coordinates. However, the result of \textit{argmax} is discrete and can only be an integer, which limits the accuracy of the final coordinates. We propose a new post-processing function named \textit{$\beta$-Soft-Argmax} to make it continuous and get more accurate predicted results. This post-processing function and the iterative training strategy above are two parameter-free compensate measures, which can improve the performance of LPN.

We empirically demonstrate the effectiveness and efficiency of our methods on the COCO \cite{lin2014microsoft} dataset. The experiments show that our LPN has much less model size and computational complexity than other top-performing networks. For example, compared with SimpleBaseline-50 (achieving 70.0 in AP score on COCO test-dev set with 34.0M parameters and 8.9 GFLOPs) \cite{xiao2018simple}, our small network LPN-50 can achieve 68.7 in AP with only 2.7M parameters and 1.0 GFLOPs. The model size (\#Params) is 9\% of SimpleBaseline-50 and the complexity (FLOPs) is 11\%, while the gap in AP is only 1.3. We also show the speed superior of our LPN-50 at inference time on a non-GPU platform. It can achieve 17 FPS on an Intel i7-8700K CPU machine.


\section{Related Work}

\subsection{Human Pose Estimation}
Traditional human pose estimation methods often follow the framework of pictorial structure model or probabilistic graphical model \cite{sapp2010cascaded,yang2011articulated,pishchulin2013poselet,pishchulin2013strong}. With the introduction of DeepPose by Toshev et al. \cite{toshev2014deeppose}, deep convolutional neural network based methods have become dominant in this area \cite{tompson2014joint,carreira2016human,lifshitz2016human,newell2016stacked,chu2017multi,chen2017adversarial}. There are two mainstream methods: regressing the position of keypoints directly \cite{toshev2014deeppose,carreira2016human}, and estimating keypoint heatmaps followed by an \textit{argmax} function to choose the maximum locations as the keypoints \cite{tompson2014joint,newell2016stacked,yang2016end}. In this work, we mainly focus on the latter.

In the last few years, significant improvements in human pose estimation have been achieved by using DCNNs. For example, Newell et al. \cite{newell2016stacked} proposed a Stacked Hourglass Network for top-down and bottom-up inference, which becomes the dominant approach on the MPII benchmark \cite{andriluka14cvpr}. He et al. \cite{he2017mask} proposed Mask R-CNN, which can perform human detection and keypoint localization in a single model. Chen et al. \cite{chen2018cascaded} proposed a Cascade Pyramid Network (CPN) to refine the process of pose estimation, which is the winner of COCO 2017 keypoint challenge. Xiao et al. \cite{xiao2018simple} provided a SimpleBaseline method that consists of a deep backbone network and several deconvolutional layers. It can achieve pretty good performance on the COCO benchmark \cite{lin2014microsoft}, although it is based on simple network architecture. Sun et al. \cite{sun2019deep} proposed a High-Resolution Network (HRNet) achieved state-of-the-art performance, which could maintain high-resolution representations by connecting high-to-low resolution convolutions in parallel and repeatedly conducting multi-scale fusions across parallel convolutions.

These prior works pay more attention to how to improve the accuracy of pose estimation by using complex architecture or computationally expensive models, ignoring the deployment costs issue in practice. The limitations such as time-consuming, high memory demanding make it extremely difficult to deploy and scale in real-world applications on edge devices.

\subsection{Lightweight Pose Estimation}
There are a few recent research for lightweight design to improve the efficiency of pose estimation networks. For example, Rafi et al. \cite{rafi2016efficient} proposed an efficient deep network architecture that can be trained efficiently on mid-range GPUs, but they didn't conduct quantitative experiments on the model efficiency. Bulat et al. \cite{bulat2017binarized} binarized the network architecture for model compression and execution speedup to accommodate resource-limited platforms, which however suffers performance drop significantly. This field is still not fully explored. These previous lightweight methods only validate their ideas on the MPII benchmark \cite{andriluka14cvpr}, which is easy and has been saturated. And these methods have limitations more or less. Following the design principles of SimpleBaseline \cite{xiao2018simple}, We present our lightweight pose network that has superiority in terms of model size, computational complexity, and inference speed. We demonstrate its efficiency and effectiveness on the more challenging COCO benchmark \cite{lin2014microsoft}.

\subsection{Attention Mechanism}
Attention mechanism has achieved great success in various computer vision tasks such as image classification \cite{wang2017residual,hu2018squeeze}, object recognition \cite{ba2014multiple}, image question answering \cite{yang2016stacked}, and so on. Chu et al. \cite{chu2017multi} firstly investigated the use of attention models for human pose estimation, proposed a method which incorporates convolutional neural networks with a multi-context attention mechanism into an end-to-end framework. For capturing long-range dependencies, Wang et al. \cite{wang2018nonlocal} proposed a Non-local network (NLNet) that adopts self-attention mechanisms to model the pixel-level pairwise relations. There is indeed some performance gain for human pose estimation, but the NLNet learns query-independent attention maps for each query position, which is a waste of computation cost.

Through rigorous empirical analysis, Cao et al. \cite{cao2019GCNet} found that the global contexts modelled by NLNet are almost the same for different query positions within an image. They designed a better instantiation, called the Global Context (GC) block, and then constructed a Global Context Network (GCNet) which can effectively model the global context via addition fusion as NLNet \cite{wang2018nonlocal}, with the lightweight property as SENet \cite{hu2018squeeze}. Hence we can apply it for each layer in our lightweight network, which can improve the network's capacity without too much computational burden increase.

\begin{figure*}[!htb]
  \centering
  \includegraphics[width=1.0\linewidth]{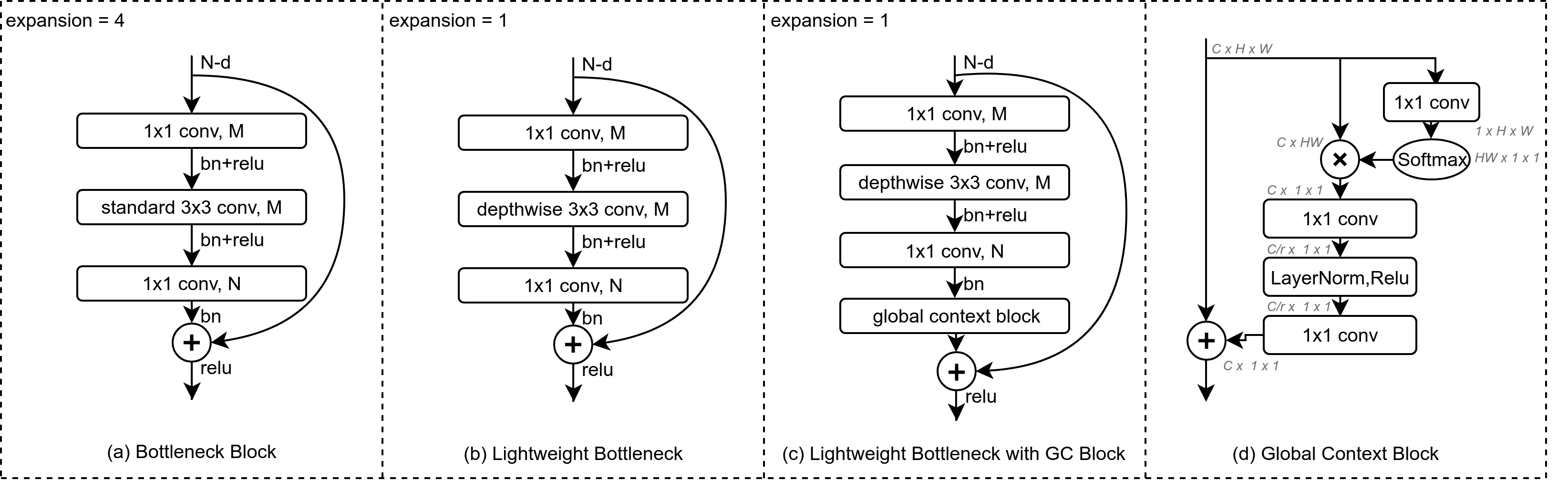}
  \caption{Architecture of the main blocks. (a) Standard Bottleneck Block in ResNet \cite{he2016deep}. (b) The redesigned Lightweight Bottleneck Block after two modifications. (c) Lightweight Bottleneck with GC Block. Note that $M$ and $N$ in these blocks denote the number of output channels of a convolutional layer. (d) Global Context Block \cite{cao2019GCNet}, which is lightweight and can effectively model long-range dependency.}
  \label{fig:bottleneck}
\end{figure*}

\section{Method}

\subsection{Lightweight Bottleneck Block}
Bottleneck block is a kind of residual block which is introduced in ResNet \cite{he2016deep}. It is the basic component of ResNet50, ResNet101, and ResNet152, the backbone networks of SimpleBaseline \cite{xiao2018simple}.

As shown in Figure {\color{red} \ref{fig:bottleneck}(a)}, a bottleneck block consists of three convolutional layers and a shortcut connection. A standard convolutional layer is parameterized by convolution kernel K of size $D_K \times D_K \times C_{in} \times C_{out}$, where $D_K$ is the spatial dimension of the kernel assumed to be square and $C_{in}$ is the number of input channels and $C_{out}$ is the number of output channels. Therefore, the total number of the parameters of a standard bottleneck block is:
\begin{equation}
1 \times 1 \times N \times M + 
3 \times 3 \times M \times M + 
1 \times 1 \times M \times N
\label{eq:standard bottleneck}
\end{equation}
Usually, for a bottleneck block, the number of input channels and output channels are consistent, both are $N$. And $N = M \times expansion$, where $M$ is the hidden dim and \textit{expansion} is a hyperparameter with default value of 4. Thus, Eqn. (\ref{eq:standard bottleneck}) can be simplified as 
\begin{equation}
  17 \times M \times M
\end{equation}

Based on two modifications of the standard bottleneck block, we present a lightweight bottleneck block. The first modification is to set the hyper-parameter \textit{expansion} to 1, which reduces the input and output channels. And the second modification is to replace the standard 3$\times$3 convolution with a 3$\times$3 \textit{depthwise convolution}, which can generate features using very few parameters. Figure {\color{red} \ref{fig:bottleneck}(b)} shows the architecture of the lightweight bottleneck block, which is similar to the standard one; nevertheless, the number of parameters has been reduced significantly, which is
\begin{equation}
  2 \times M \times M + 3 \times 3 \times M
\end{equation}
after simplify. Thus, we can get a reduction in parameters of
\begin{equation}
  \frac{2 \times M \times M + 3 \times 3 \times M} {17 \times M \times M } \approx \frac{2}{17}
\end{equation}
The reduction in computational costs is similar to the reduction in parameters, since they are proportional to each other.

We also attempt to equip the lightweight block with a global context (GC) block \cite{cao2019GCNet}, which can capture long-range dependencies while keeping lightweight and efficient. The concept of long-range dependencies has been empirically proven to be effective to human pose estimation in Non-local network \cite{wang2018nonlocal}. Figure {\color{red} \ref{fig:bottleneck}(c)} and {\color{red} \ref{fig:bottleneck}(d)} show the architecture of the lightweight bottleneck block with GC block and the details of GC block, respectively. The use of GC block can improve the capacity of the lightweight bottleneck block without too much computational burden increase. Experiments show that the GC block is more friendly to small network.

\subsection{Lightweight Pose Network}
\label{section:LPN}
The widely-adopted pipeline \cite{newell2016stacked,chen2018cascaded,xiao2018simple,sun2019deep} to predict human keypoints is composed of a stem decreasing the resolution, a main body outputting the feature maps, and a regressor estimating the heatmaps where the keypoint positions are chosen and transformed to the full resolution. 

SimpleBaseline \cite{xiao2018simple} uses a ResNet backbone as the main body, and uses three deconvolutional layers as the regressor. Following its design principles, we present a Lightweight Pose Network (LPN). Different with SimpleBaseline, we replace the standard bottleneck blocks used in the backbone with our lightweight bottleneck blocks when downsampling, which drastically reduces the scale of parameters and FLOPs. During the upsampling process, we replace each deconvolutional layer with a combination of a group deconvolutional layer and a 1$\times$1 convolutional layer to reduce the redundant parameters while keeping the quality of upsampling. For convenience, we set the group size of the group deconvolutions to the greatest common divisor of input channels and output channels. 

Intuitively, it is beneficial to maintain high-resolution representations before upsampling. HRNet \cite{sun2019deep} maintains high-resolution representations through the whole process, which empirically demonstrate the idea above by the superior pose estimation results. However, most existing methods tend to produce low-resolution representations (e.g., $6 \times 8$, $4 \times 4$) probably because they scruple that higher-resolution representations will increase the computational burden heavily. In order to keep the characteristics of lightweight while obtaining higher-resolution representations, we remove the downsampling in layer4\footnote{The layer4 here means a sequence of lightweight bottleneck blocks, not an individual convolutional layer.} and delete a group deconvolutional layer at the same time. The architecture of LPN is illustrated in Figure {\color{red} \ref{fig:network}}. According to our design, the sizes of feature maps of the last two stages are the same in the backbone network, and there are only two group deconvolutional layers (GD1, GD2) in the upsampling process.

\subsection{Iterative Training Strategy}

We present the iterative training strategy for two reasons: 
\begin{itemize}
  \item {
    Human pose networks rely heavily on pre-trained models. It is not advisable to train the network on ImageNet to get a pre-trained model because of the time costs problem. Our iterative training strategy can overcome this barrier, providing a pre-trained model in a different way while saving time costs.}
  \item {
    Such an unconventional training strategy can give full play to LPN's potential. The famous \textit{universal approximation theorem} \cite{hornik1989multilayer,cybenko1989approximation} means that a network will be able to represent regardless of what function, provided that the network is given enough hidden units. That is to say, the network has the ability to reach the global minima given enough parameters, but in fact, it always falls into some local minima during optimization. Small networks may be easier to fall into local minima because of the weak generalization capability. The smaller, the easier. Once falling into local minima, it is difficult for the optimizer with a small learning rate to cross the ridge. Therefore, sometimes it is necessary to increase the learning rate again.
  }
\end{itemize}

\begin{figure}[t]
  \centering
  \includegraphics[width=\linewidth]{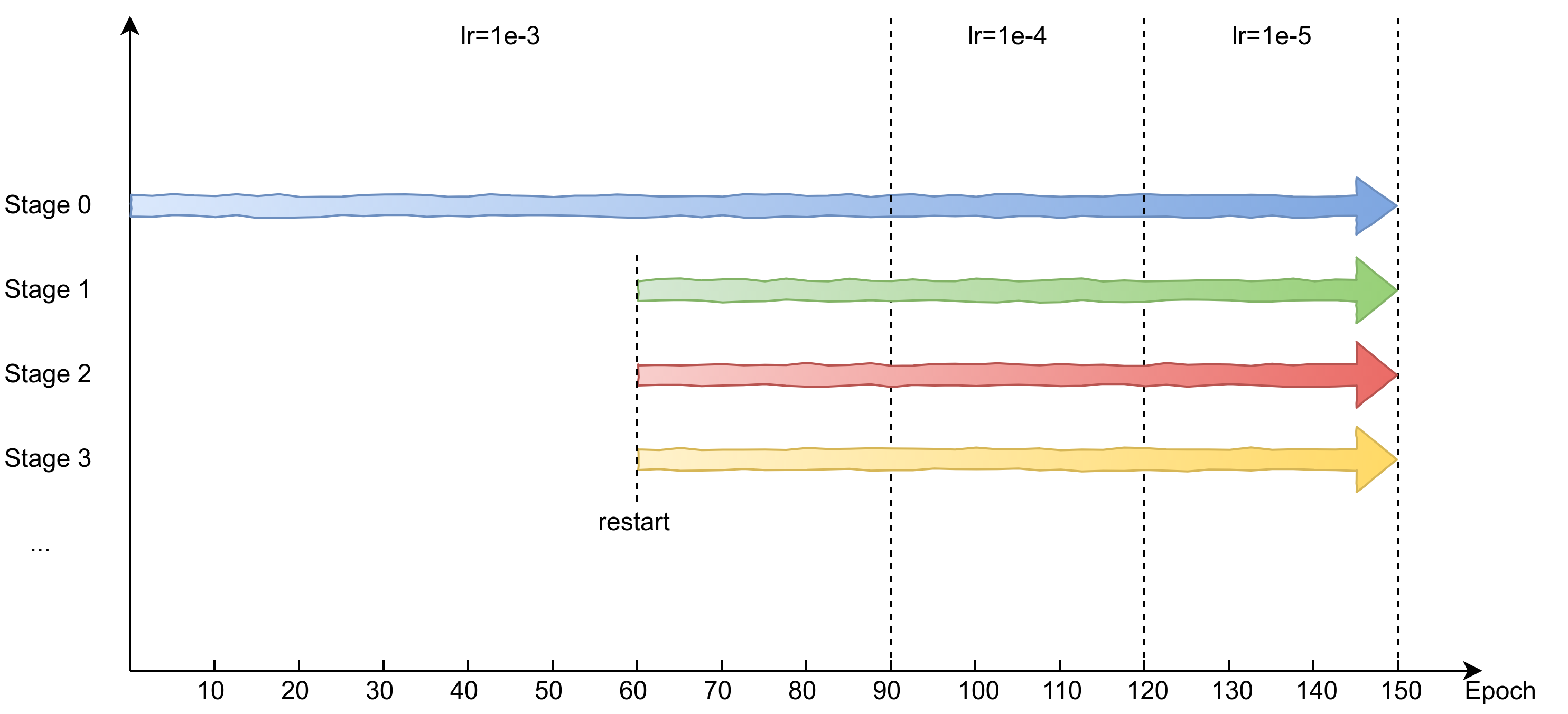}
  \caption{Illustrating the iterative training strategy.}
  \label{fig:iterative}
\end{figure}

Our iterative training strategy mainly focuses on the learning rate during training. Almost all existing methods follow the same training pipeline, where they first initialize the learning rate with a specific value (e.g., 1e-3), and then decay the learning rate every time the number of epochs reaches one of the milestones. Differently, our training strategy changes the learning rate periodically. It can be divided into multiple stages: the first stage is the same as the pipeline mentioned above; at each subsequent stage, use the best model obtained in the previous stage as a pre-trained model to initialize the parameters, reset the learning rate, restart training from a specified epoch, and repeat the rest process of the pipeline. The entire training process is illustrated in Figure {\color{red} \ref{fig:iterative}}, which is also the config details about our experiments.

The ablation study shows that our iterative training strategy, with learning rate changing periodically, is very effective for LPN. The performance of LPN spirals upwards until it converges. The improvement in score is more significant than using a pre-trained model. Last but not least, there is no need to spend any more energy on getting a pre-trained model.

\subsection{$\beta$-Soft-Argmax}

At inference time, most existing methods use function \textit{argmax} to get the keypoint positions in heatmaps and transform to the full resolution. The result of \textit{argmax} is discrete and can only be an integer, which limits the accuracy of the final coordinates. Luvizon et al. \cite{luvizon2019human} and Sun et al. \cite{sun2018integral} have attempted to use a technique named \textit{Soft-Argmax} to regress the final coordinates, making the whole process differentiable. However, we notice that the ground-truth heatmaps are generated by applying 2D Gaussian centered on each keypoint
\begin{equation}
  H_k(x,y)=\exp(-\frac{(x-x_k)^2+(y-y_k)^2}{2\sigma^2})
\end{equation}  
where $H_k$ and $(x_k,y_k)$ are the ground-truth heatmap and the 2D coordinates of $k$th keypoint, respectively, and $\sigma$ is the standard deviation of the Gaussian peak. The ground truth heatmaps normalized to $[0,1]$, means that there will be a large number of values close to 0 in the predicted heatmaps, which may affect the accuracy of \textit{Soft-Argmax} when applying spatial softmax through
\begin{equation}
  S_k(x,y) = \frac{e^{H_k(x,y)}}{\sum_x\sum_y e^{H_k(x,y)}}
\end{equation}
Since $e^0=1, e^1=e$, a lot of zeros in heatmaps will reduce the probability of the maximum and then affect the accuracy of the results.

We present a method named \textit{$\beta$-Soft-Argmax}, where we optimize the formula of \textit{Soft-Argmax} by adding a coefficient $\beta$ before the heatmap $H_k(x,y)$ to suppress the effect of the values close to zero. 
\begin{equation}
  S_k(x,y) = \frac{e^{\beta H_k(x,y)}}{\sum_x\sum_y e^{\beta H_k(x,y)}}, \ \ (\beta > 1)
\end{equation}
Same as \textit{Soft-Argmax}, the final coordinates are obtained through 
$\widehat{x}_k = \sum S_k \circ W_x, \ \widehat{y}_k = \sum S_k \circ W_y\ $, where $W_x$ and $W_y$ are constant weight matrixes and $\circ$ means element-wise multiplication. We show that our \textit{$\beta$-Soft-Argmax} is a model-agnostic general function in the post process, it is beneficial not only for our LPN, but also for existing networks, such as SimpleBaseline \cite{xiao2018simple}, HRNet \cite{sun2019deep}.

\section{Experiments}

\begin{table*}[t]
  \footnotesize
  \centering
  \caption{Comparisons of results on COCO validation set. Pretrain = pretrain the backbone on the ImageNet classification task.}
  \vspace{-0.1cm}
  \begin{spacing}{1.2}
  \resizebox{\linewidth}{!}{
    \begin{tabular}{|l|l|c|c|r|r|cccccc|}
      \hline
      Method & Backbone & Pretrain & Input size & \#Params & FLOPs & $\operatorname{AP}$ & $\operatorname{AP}^{50}$ & $\operatorname{AP}^{75}$ & $\operatorname{AP}^{M}$ & $\operatorname{AP}^{L}$ & $\operatorname{AR}$ \\

      \hline
      $8$-stage Hourglass~\cite{newell2016stacked} 
      & Hourglass & N & $256\times192$ & $25.6$M & $26.2$G 
      & $66.9$ & $-$ & $-$ & $-$ & $-$ & $-$ \\ 
  
      CPN~\cite{chen2018cascaded} 
      & ResNet-50 & Y & $256\times192$ & $27.0$M & $6.2$G 
      & $68.6$ & $-$ & $-$ & $-$ & $-$ & $-$ \\ 
    
      SimpleBaseline~\cite{xiao2018simple} 
      & ResNet-50 & Y & $256\times192$ & $34.0$M & $8.9$G 
      & $70.4$ & $88.6$ & $78.3$ & $67.1$ & $77.2$ & $76.3$ \\
    
      SimpleBaseline~\cite{xiao2018simple} 
      & ResNet-101 & Y & $256\times192$ & $53.0$M & $12.4$G 
      & $71.4$ & $89.3$ & $79.3$ & $68.1$ & $78.1$ & $77.1$ \\
    
      SimpleBaseline~\cite{xiao2018simple} 
      & ResNet-152 & Y & $256\times192$ & $68.6$M & $15.7$G 
      & $72.0$ & $89.3$ & $79.8$ & $68.7$ & $78.9$ & $77.8$ \\

      HRNet-W$32$~\cite{sun2019deep} 
      & HRNet-W$32$ & N & $256\times192$& $28.5$M & $7.1$G 
      & $73.4$ & $89.5$ & $80.7$ & $70.2$ & $80.1$ & $78.9$ \\

      HRNet-W$32$~\cite{sun2019deep} 
      & HRNet-W$32$ & Y & $256\times192$ & $28.5$M & $7.1$G 
      & $74.4$ & $90.5$ & $81.9$ & $70.8$ & $81.0$ & $79.8$ \\

      HRNet-W$48$~\cite{sun2019deep} 
      & HRNet-W$48$ & Y & $256\times192$ & $63.6$M & $14.6$G 
      & $75.1$ & $90.6$ & $82.2$ & $71.5$ & $81.8$ & $80.4$ \\
  
      \hline
      SimpleBaseline~\cite{xiao2018simple} 
      & ResNet-152 & Y & $384\times288$ & $68.6$M & $35.6$G 
      & $74.3$ & $89.6$ & $81.1$ & $70.5$ & $79.7$ & $79.7$ \\

      HRNet-W$48$~\cite{sun2019deep} 
      & HRNet-W$48$ & Y & $384\times288$ & $63.6$M & $32.9$G 
      & $\textbf{76.3}$ & $\textbf{90.8}$ & $\textbf{82.9}$ &$\textbf{72.3}$ & $\textbf{83.4}$ & $\textbf{81.2}$\\

      \hline
      LPN (Ours) 
      & ResNet-50 & N & $256\times192$ & $\textbf{2.9}$M & $\textbf{1.0}$G 
      & $69.1$ & $88.1$ & $76.6$ & $65.9$ & $75.7$ & $74.9$ \\

      LPN (Ours) 
      & ResNet-101 & N & $256\times192$ & $5.3$M & $1.4$G 
      & $70.4$ & $88.6$ & $78.1$ & $67.2$ & $77.2$ & $76.2$\\

      LPN (Ours) 
      & ResNet-152 & N & $256\times192$ & $7.4$M & $1.8$G 
      & $71.0$ & $89.2$ & $78.6$ & $67.8$ & $77.7$ & $76.8$\\

      \hline
    \end{tabular}
  }
  \end{spacing}

  \label{table:coco_val}
\end{table*}

\begin{table*}[t]
  \footnotesize
  \centering
  \caption{Comparisons of results on COCO test-dev set. \#Params and FLOPs are calculated for the pose estimation network, and those for human detection are not included. M = $10^6$, G = $2^{30}$.}
  \vspace{-0.1cm}
  \begin{spacing}{1.2}
  \resizebox{\linewidth}{!}{
    \begin{tabular}{|l|l|c|r|r|cccccc|}
      \hline
      Method & Backbone & Input size & \#Params & FLOPs & $\operatorname{AP}$ & $\operatorname{AP}^{50}$ & $\operatorname{AP}^{75}$ & $\operatorname{AP}^{M}$ & $\operatorname{AP}^{L}$ & $\operatorname{AR}$ \\
      
      \hline
      Mask-RCNN~\cite{he2017mask} 
      & ResNet-50-FPN & $-$ & $-$ & $-$ 
      & $63.1$ & $87.3$ & $68.7$ & $57.8$ & $71.4$ & $-$ \\

      G-RMI~\cite{papandreou2017towards} 
      & ResNet-101 & $353\times257$ & $42.6$M & $57.0$G 
      & $64.9$ & $85.5$ & $71.3$ & $62.3$ & $70.0$ & $69.7$ \\

      Integral Regression~\cite{sun2018integral} 
      & ResNet-101 & $256\times256$ & $45.0$M & $11.0$G 
      & $67.8$ & $88.2$ & $74.8$ & $63.9$ & $74.0$ & $-$ \\

      SimpleBaseline~\cite{xiao2018simple} 
      & ResNet-50 & $256\times192$ & $34.0$M & $8.9$G 
      & $70.0$ & $90.9$ & $77.9$ & $66.8$ & $75.8$ & $75.6$ \\

      SimpleBaseline~\cite{xiao2018simple} 
      & ResNet-152 & $256\times192$ & $68.6$M & $15.7$G 
      & $71.6$ & $91.2$ & $80.1$ & $68.7$ & $77.2$ & $77.3$ \\

      CPN~\cite{chen2018cascaded} 
      & ResNet-Inception & $384\times288$ & $-$ & $-$ 
      & $72.1$ & $91.4$ & $80.0$ & $68.7$ & $77.2$ & $78.5$ \\

      RMPE~\cite{fang2017rmpe} 
      & PyraNet~\cite{yang2017learning} & $320\times256$ & $28.1$M & $26.7$G 
      & $72.3$ & $89.2$ & $79.1$ & $68.0$ & $78.6$ & $-$ \\

      SimpleBaseline~\cite{xiao2018simple} 
      & ResNet-152 & $384\times288$ & $68.6$M & $35.6$G 
      & $73.7$ & $91.9$ & $81.1$ & $70.3$ & $80.0$ & $79.0$ \\

      HRNet-W$32$~\cite{sun2019deep} 
      & HRNet-W$32$ & $384\times288$ & $28.5$M & $16.0$G 
      & $74.9$ & $92.5$ & $82.8$ & $71.3$ & $80.9$ & $80.1$ \\

      HRNet-W$48$~\cite{sun2019deep} 
      & HRNet-W$48$ & $384\times288$ & $63.6$M & $32.9$G 
      & $\textbf{75.5}$ & $\textbf{92.5}$ & $\textbf{83.3}$ & $\textbf{71.9}$ & $\textbf{81.5}$ & $\textbf{80.5}$ \\

      \hline
      LPN (Ours) 
      & ResNet-50 & $256\times192$ & $\textbf{2.9}$M & $\textbf{1.0}$G 
      & $68.7$ & $90.2$ & $76.9$ & $65.9$ & $74.3$ & $74.5$ \\

      LPN (Ours) 
      & ResNet-101 & $256\times192$ & $5.3$M & $1.4$G 
      & $70.0$ & $90.8$ & $78.4$ & $67.2$ & $75.4$ & $75.7$\\

      LPN (Ours) 
      & ResNet-152 & $256\times192$ & $7.4$M & $1.8$G 
      & $70.4$ & $91.0$ & $78.9$ & $67.7$ & $76.0$ & $76.2$\\
  
      \hline
    \end{tabular}
  }
  \end{spacing}

  \label{table:coco_test_dev}
\end{table*}

\subsection{COCO Keypoint Detection}
\label{section:experiment}

\textbf{Dataset and evaluation metric.}
The COCO dataset \cite{lin2014microsoft} contains over 200K images and 250K person instances labeled with 17 keypoints. We train our lightweight network on the train2017 set, including 57K images and 150K person instances. We evaluate our method on the val2017 set and test-dev2017 set, containing 5K images and 20K images, respectively. The OKS-based AP metric is used to evaluate the accuracy of the keypoint localization.

\vspace{0.2cm}
\textbf{Training.}
The presented LPN is trained in an end-to-end manner. All parameters are initialized randomly from zero-mean Gaussian distribution with $\sigma=0.001$. We use Adam optimizer with a mini-batch size of 32 to update the parameters. The initial learning rate is set to 1e-3 and reduced by a factor of 10 at the 90th and 120th epoch. We apply our iterative training strategy, which is illustrated in Figure {\color{red} \ref{fig:iterative}}. In each retraining stage, we initialize the parameters with the model generated in the previous stage, then we reset the learning rate to 1e-3 and restart at the 60th epoch. The training process of each stage is terminated at the 150th epoch. There are 7 stages in total, including the first one.

In terms of data processing, same as \cite{xiao2018simple}, we extend the human detection box to a fixed aspect ratio (e.g., height: width = 4:3), and then crop the box from the image. The cropped bounding box is resized to a fixed size (256$\times$192), which becomes the input image. The data augmentation includes random rotation ($\pm 40^\circ$), random scale ($\pm 30 \%$), and flipping. All of our experiments are performed on an NVIDIA 1080Ti GPU.

\vspace{0.2cm}
\textbf{Testing.}
The two-stage top-down paradigm is used: detect the person instance using a person detector, and then predict keypoints. We use the same person detectors provided by SimpleBaseline\footnote{https://github.com/Microsoft/human-pose-estimation.pytorch} \cite{xiao2018simple} for both validation set and test-dev set. Following the common practice \cite{newell2016stacked,sun2019deep}, we compute the heatmaps by averaging the heatmaps of the original and flipped images. Each keypoint location is obtained by using \textit{$\beta$-Soft-Argmax} and inverse affine transformation. The hyperparameter $\beta$ is set to 160 at inference time.

\vspace{0.2cm}
\textbf{Results on the validation set.}
As shown in Table {\color{red} \ref{table:coco_val}}, we have 3 different versions of the lightweight pose network (LPN-50, LPN-101, LPN-152) according to the depth of the backbone, which is similar to SimpleBaseline \cite{xiao2018simple}. Our smallest network LPN-50 achieves 69.1 in AP score, only with 2.9M parameters and 1.0 GFLOPs.
(\romannumeral1) Compared to Hourglass \cite{newell2016stacked} and CPN \cite{chen2018cascaded}, LPN-50 outperforms both of them, while the model size is considerably smaller and the computational complexity is much lower than them.
(\romannumeral2) Compared to SimpleBaseline \cite{xiao2018simple}, there is a little gap (1.3 points) between LPN-50 and the corresponding one (Simple-50). However, the number of the parameters and the FLOPs of LPN-50 are only 9\% and 11\% of Simple-50, respectively.
(\romannumeral3) Compared to the best-performed HRNet \cite{sun2019deep}, although they can achieve a higher result in AP score, the speed at inference time is much lower than our lightweight network, because there are lots of parallel convolutions in their architecture. See Section \ref{section:speed} for details. Besides, our network is also advantageous in terms of deployment because of the smaller model size.

Our network can benefit from increasing the depth of the backbone, 0.7 and 1.9 improvements for LPN-101 and LPN-152, respectively. 
The results also benefit from our iterative training strategy, which has a better effect than using a pre-trained model so that there is no need to pre-train on the ImageNet classification task, and time costs saved.

\vspace{0.2cm}
\textbf{Results on the test-dev set.}
Table {\color{red} \ref{table:coco_test_dev}} reports the pose estimation performance of our method and the existing state-of-the-art methods. The results of SimpleBaseline \cite{xiao2018simple} with input size 256$\times$192 are obtained using the official model files\footnotemark[3]. We don't increase the input size to pursue higher scores like other methods. Our networks, achieving acceptable results, are the most efficient in terms of model size (\#Params) and computational complexity (FLOPs).

\subsection{Ablation Study}
\label{section:ablation}

We study the effect of each component in our methods, including the global context block, the iterative training strategy, and the post-processing function \textit{$\beta$-Soft-Argmax}. We validate our methods on the COCO val2017 set, all results are obtained over the input size of $256\times192$.

\vspace{0.2cm}
\textbf{GC block.} 
We study the effect of GC block \cite{cao2019GCNet}. We build our networks using the lightweight bottleneck without and with GC block, respectively and train them. In this part, we only apply the first stage of the iterative training strategy without repeated training. The post-processing function \textit{$\beta$-Soft-Argmax} is not used too. The AP scores of the networks are reported in Table {\color{red} \ref{table:gc_block}}. We can find that the improvement of GC block on the small network is more obvious than on the large network, e.g., the improvement is 2.5 points for LPN-50, 1.1 points for LPN-101, and 0.4 points for LPN-152. GC block does improve the prediction accuracy, but with the increase of the network's depth, the gain will decrease gradually. It is more friendly to small networks. And the increase in both \#Params and FLOPs is within the acceptable range.

\vspace{0.2cm}
\textbf{Iterative training strategy.}
We study how the iterative training strategy affects the pose estimation performance step by step. We train our lightweight networks following the descriptions in Section \ref{section:experiment} and evaluate the performance when each training stage ends. 

Results in Table {\color{red} \ref{table:coco_val}} show that, for HRNet-W32 \cite{sun2019deep}, the gain is 1.0 points when training from the model pre-trained for the ImageNet classification problem. We report the gain of our iterative training strategy in Table {\color{red} \ref{table:iterative}}. As the number of stages increases, the performance of our LPN is improved until converges. The table shows that the cumulative gain of our iterative training strategy is more significant than using a pre-trained model, e.g., the improvement is 2.02 points for LPN-50, 1.40 points for LPN-101, and 1.50 points for LPN-152. The excellent improvement also implies that lightweight networks are hard to train, perhaps because they are easier to fall into the local minima. Our iterative training strategy, with learning rate changing periodically, can give full play to LPNs' potential.

\begin{table}
  \footnotesize
  \centering
  \caption{Ablation study of GC block. $\times$ and $\surd$ correspond to using the lightweight bottleneck without and with GC block, respectively. }
  \vspace{-0.1cm}
  \begin{spacing}{1.2}
  \resizebox{\linewidth}{!}{
    \begin{tabular}{l|c|r|r|cccc}
      \hline
      Method & GC Block & \#Params & FLOPs & $\operatorname{AP}$ & $\operatorname{AP}^{50}$ & $\operatorname{AR}$ \\

      \hline
      LPN-50 & $\times$ & $2.7$M & $1.0063$G 
      & $64.4$ & $86.1$ & $70.6$ \\
      LPN-50 & $\surd$ & $2.9$M & $1.0079$G 
      & $66.9$ & $87.2$ & $73.0$ \\

      \hline
      LPN-101 & $\times$ & $5.0$M & $1.4118$G 
      & $67.8$ & $87.5$ & $73.8$ \\
      LPN-101 & $\surd$ & $5.3$M & $1.4143$G 
      & $68.9$ & $88.2$ & $74.8$ \\

      \hline
      LPN-152 & $\times$ & $6.9$M & $1.8189$G 
      & $69.0$ & $88.0$ & $74.9$ \\
      LPN-152 & $\surd$ & $7.4$M & $1.8224$G 
      & $69.4$ & $88.6$ & $75.3$ \\

      \hline
    \end{tabular}
  }
  \end{spacing}
  \vspace{-0.5cm}
  \label{table:gc_block}
\end{table}

\vspace{0.2cm}
\textbf{$\beta$-Soft-Argmax.}
At last, we study the effect of our post-processing function \textit{$\beta$-Soft-Argmax}. We evaluate the impact of $\beta$ with different values on our LPNs and the top-performing networks (e.g., SimpleBaseline \cite{xiao2018simple}, HRNet \cite{sun2019deep}). Our LPNs are trained using the iterative strategy, and those top-performing networks are initialized using the official trained model files\footnote{https://github.com/leoxiaobin/deep-high-resolution-net.pytorch}. 

The comparisons of results are shown in Table {\color{red} \ref{table:beta}}. We can find that it is unhelpful or even harmful when $\beta$ is small, but the gain becomes obvious gradually with the increase of $\beta$. When $\beta \ge 120$, the improvement tends to be saturated. We choose 160 as the value of $\beta$ finally. The most significant improvement is 0.3 points in AP when applied on Simple-152 \cite{xiao2018simple}, others float between 0.1 and 0.2. Although the gain of function \textit{$\beta$-Soft-Argmax} is not so significant as the iterative training strategy, it can be a new model-agnostic post-processing function at inference time to pursue more accurate predictions.

\begin{table}
  \footnotesize
  \centering
  \caption{Ablation study of the iterative training strategy. Values in the last row are the cumulative gain in AP score.}
  \vspace{-0.25cm}
  \begin{spacing}{1.2}
  \resizebox{0.9\linewidth}{!}{
    \begin{tabular}{|c|c|c|c|c|c|c|}
      \hline
      \multirow{2}{*}{} & \multicolumn{2}{c|}{LPN-50} & \multicolumn{2}{c|}{LPN-101} & \multicolumn{2}{c|}{LPN-152} \\ 
      \cline{2-7} & $\operatorname{AP}$ & $\Delta$gain & $\operatorname{AP}$ & $\Delta$gain & $\operatorname{AP}$ & $\Delta$gain \\ 
      \hline
      Stage 0 & 66.90 &   -   & 68.85 &   -   & 69.36 &   -   \\ 
      \hline
      Stage 1 & 67.73 & +0.83 & 69.34 & +0.49 & 69.87 & +0.51 \\ 
      \hline
      Stage 2 & 68.12 & +0.39 & 69.61 & +0.27 & 70.27 & +0.40 \\ 
      \hline
      Stage 3 & 68.28 & +0.16 & 69.78 & +0.17 & 70.57 & +0.30 \\ 
      \hline
      Stage 4 & 68.69 & +0.41 & 69.95 & +0.17 & 70.68 & +0.11 \\ 
      \hline
      Stage 5 & 68.89 & +0.20 & 70.15 & +0.20 & 70.86 & +0.18 \\ 
      \hline
      Stage 6 & 68.92 & +0.03 & 70.25 & +0.10 & 70.80 & -0.06 \\ 
      \hline
              &       & +2.02 &       & +1.40 &       & +1.50 \\ 
      \hline
    \end{tabular}
  }
  \end{spacing}

  \label{table:iterative}
\end{table}

\begin{table}
  \footnotesize
  \centering
  \caption{Ablation study of \textit{$\beta$-Soft-Argmax}. $Argmax$ denotes using function \textit{Argmax} at inference time, and $\beta=[20:200]$ denotes using our \textit{$\beta$-Soft-Argmax} with different values.}
  \vspace{-0.25cm}
  \begin{spacing}{1.2}
  \resizebox{0.9\linewidth}{!}{
    \begin{tabular}{|l|cccccccc|}
      \hline
      \multicolumn{1}{|l|}{} & \multicolumn{8}{c|}{AP(\%)} \\
      \hline
      Method & $Argmax$ & $\beta=20$ & $\beta=40$ & $\beta=60$ & $\beta=80$ & $\beta=120$ & $\beta=160$ & $\beta=200$\\ 
      \hline
      Simple-50~\cite{xiao2018simple} 
      & 70.4 & 65.5 & 69.6 & 70.2 & 70.4 & 70.6 & \textbf{70.6} & 70.6 \\
      Simple-101~\cite{xiao2018simple} 
      & 71.4 & 67.2 & 70.6 & 71.3 & 71.5 & 71.5 & \textbf{71.5} & 71.5 \\
      Simple-152~\cite{xiao2018simple}
      & 72.0 & 67.9 & 71.4 & 72.0 & 72.2 & 72.3 & \textbf{72.3} & 72.2\\
      \hline
      HRNet-W32~\cite{sun2019deep} 
      & 74.4 & 70.8 & 73.9 & 74.4 & 74.6 & 74.5 & \textbf{74.6} & 74.5\\
      HRNet-W48~\cite{sun2019deep} 
      & 75.0 & 72.2 & 74.6 & 75.1 & 75.2 & 75.2 & \textbf{75.2} & 75.2\\
      \hline
      LPN-50 
      & 68.9 & 63.1 & 67.9 & 68.6 & 68.9 & 69.1 & \textbf{69.1} & 69.1\\
      LPN-101 
      & 70.3 & 65.1 & 69.4 & 70.1 & 70.3 & 70.4 & \textbf{70.4} & 70.4\\
      LPN-152 
      & 70.9 & 65.9 & 70.0 & 70.7 & 70.9 & 71.0 & \textbf{71.0} & 71.0\\
      \hline
    \end{tabular}
  }
  \end{spacing}

  \label{table:beta}
\end{table}

\subsection{Inference Speed}
\label{section:speed}
\begin{figure}[t]
  \centering
  \includegraphics[width=1.0\linewidth]{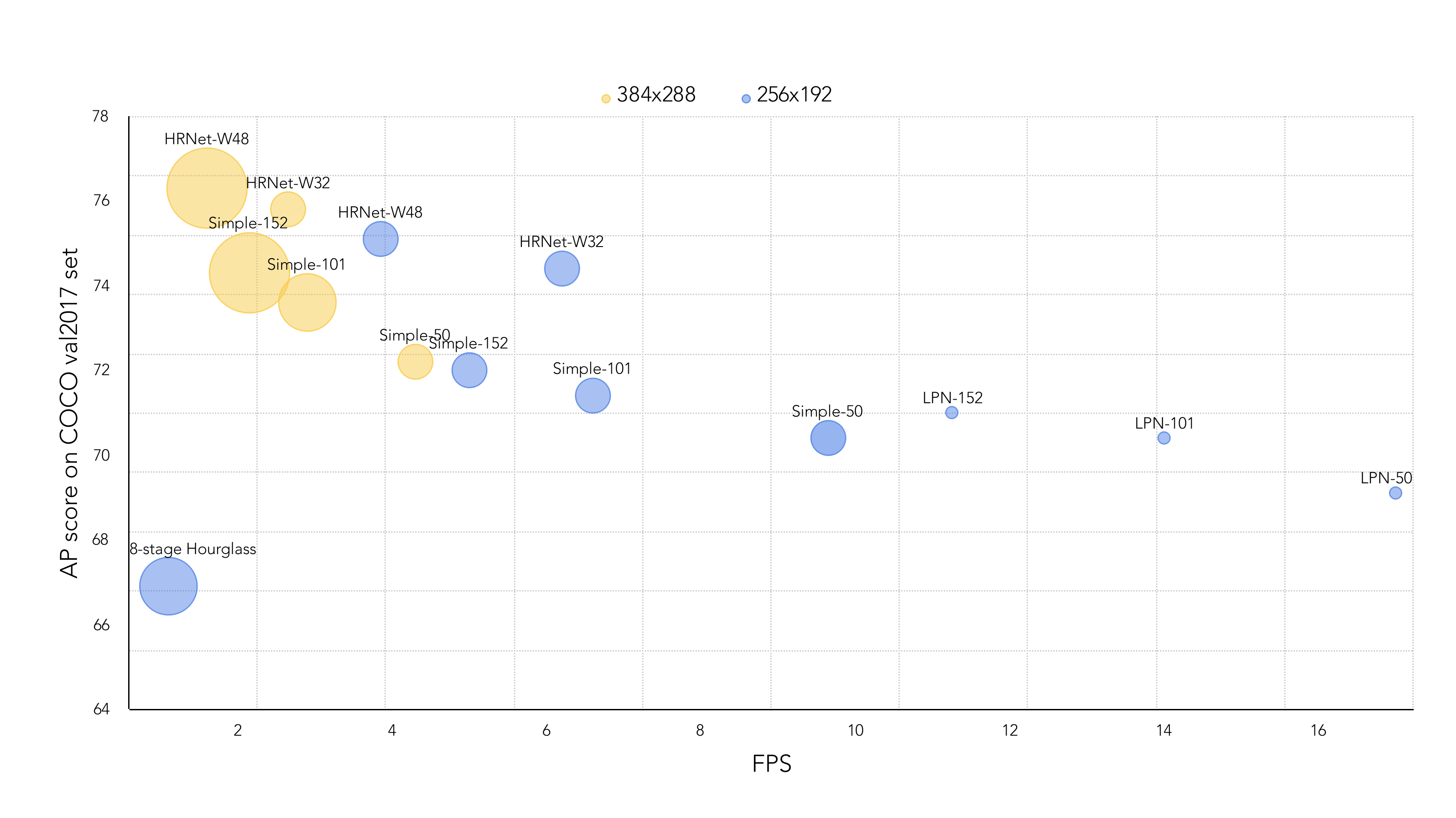}
  \caption{Measurement of AP score, speed and FLOPs of the network architecture referred in Table \ref{table:coco_val} on a non-GPU platform. Two different colors denote different input sizes, $384\times288$ and $256\times192$. The area of a circle represents the scale of the FLOPs of the corresponding method.} 
  \label{fig:speed}
\end{figure}

FLOPs, the number of float-point operations, is a widely used metric when measuring the computational complexity of a network. However, it is not a direct metric that we really care about, such as the running speed. In this section, we study the actual inference speed of our LPNs and some popular human pose estimation networks on a non-GPU platform. The platform is based on Intel Core i7-8700K CPU (3.70GHz$\times$12). Figure {\color{red} \ref{fig:speed}} shows the measurement of AP score, inference speed and FLOPs of the network architecture referred in Table \ref{table:coco_val}. 

We can find that a network doesn't necessarily run faster even if it has fewer FLOPs. For example, the FLOPs of Simple-152 \cite{xiao2018simple} is slightly more than HRNet-W48 \cite{sun2019deep}, 15.7 GFLOPs for Simple-152(blue) and 14.6 GFLOPs for HRNet-W48(blue), but the inference speed (FPS) of Simple-152 is faster than HRNet-W48, because there are lots of parallel convolutions in HRNet. We can also find that a larger input size indeed improve the predicted accuracy, however, the cost can not be ignored. For example, the AP improvement is 1.2 points for HRNet-W48 when increasing the input size, while its inference speed decays from 3.8 FPS to 1.6 FPS.

Our small network LPN-50 can achieve 17 FPS on the non-GPU platform, which is about 3 times the inference speed of Simple-152 \cite{xiao2018simple} and HRNet-W32 \cite{sun2019deep}. And our large network LPN-152 can achieve 11 FPS while keeping an AP score of 71.0. Therefore, our lightweight pose networks are demonstrated much more efficient than the top-performing architecture in theory and practice. It implies that our networks can better meet the needs of practical applications on edge devices. 

\section{Conclusion}
In this paper, we present a simple and lightweight network for human pose estimation, an iterative training strategy for better performance, and a post-processing function \textit{$\beta$-Soft-Argmax} for more accurate predictions. Our methods can achieve pretty decent results on the COCO dataset when compared with those top-performing methods, but our network is much more efficient than them in terms of inference speed. We hope our methods will be helpful when somebody develops their lightweight models. And we also hope our methods could inspire more idea on the lightweight human pose estimation field.

\section{Acknowledgement}
The authors thank Limin Wang for helpful discussions.

\newpage
\bibliography{lightweight-pose}
\end{document}